# Efficient dynamic point cloud coding using Slice-Wise Segmentation

*Faranak Tohidi[1], Manoranjan Paul[1], Anwaar Ulhaq[2]*
1 Charles Sturt University, Bathurst, Australia
2 Charles Sturt University, Port Macquarie, Australia
{ftohidi,mpaul,aulhaq}@csu.edu.au

*Abstract*— With the fast growth of immersive video sequences, achieving seamless and high-quality compressed 3D content is even more critical. MPEG recently developed a video-based point cloud compression (V-PCC) standard for dynamic point cloud coding. However, reconstructed point clouds using V-PCC suffer from different artifacts, including losing data during pre-processing before applying existing video coding techniques, e.g., High-Efficiency Video Coding (HEVC). Patch generations and self-occluded points in the 3D to the 2D projection are the main reasons for missing data using V-PCC. This paper proposes a new method that introduces overlapping slicing as an alternative to patch generation to decrease the number of patches generated and the amount of data lost. In the proposed method, the entire point cloud has been cross-sectioned into variable-sized slices based on the number of self-occluded points so that data loss can be minimized in the patch generation process and projection. For this, a variable number of layers are considered, partially overlapped to retain the self-occluded points. The proposed method's added advantage is to reduce the bits requirement and to encode geometric data using the slicing base position. The experimental results show that the proposed method is much more flexible than the standard V-PCC method, improves the rate-distortion performance, and decreases the data loss significantly compared to the standard V-PCC method.

## I. INTRODUCTION

Realistic digital representations of 3D objects and surroundings have been recently made possible. This is due to recent advances in computer graphics allowing real-time and realistic physical world interactions with users [1,2]. Emerging technologies enable real-world objects, persons, and scenes to move dynamically across users' views convincingly using a 3D point cloud [3-5]. A point cloud is a set of individual 3D points that are not organized and without any relationship in the 3D space [1,6]. Each point has a 3D position but can also contain some other attributes (e.g., texture, reflectance, colour, and normal), creating a realistic visual representation model for static and dynamic 3D objects [3,7]. This is desirable for many applications such as geographic information systems, cultural heritage, immersive telepresence, telehealth, disabled access, 3D telepresence, telecommunication, autonomous driving, gaming and robotics, virtual reality (VR), and augmented reality (AR) [2,8]. Even the use of point cloud in Metaverse when creating an avatar or content in Metaverse and object-based interaction is required. The Metaverse is a virtual world that creates a network where anyone can interact through their avatars [9]. Therefore, it is critical to present the 3D virtual world as close to the real world as possible, with high-resolution and minimal noise and blur.

Point clouds with high density (up to millions of 3D points) are required for hyper-realistic visualization applications, which raise enormous issues regarding computational and memory requirements [1,10]. Therefore, creating 3D high-fidelity content requires enormous storage, transmission, processing, and visualization resources. Several compression methods have been tried recently, and none is the complete solution. However, two main approaches were standardized in 2020 and early 2021, which are video-based point cloud compression (V-PCC) and geometry-based point cloud compression (G-PCC) [1-3]. The data structure used in G-PCC is great for static scenes. Therefore, it can address the point cloud compression in both Category 1 (static point clouds) and Category 3 (dynamically acquired point clouds) [11-13]. V-PCC is used for Category 2 (dynamic point clouds) due to its better performance in compressing dynamic scenes than G-PCC [11, 14]. However, the artifacts produced by V-PCC, especially in low bitrate, could affect user experience [15]. Losing data during pre-processing and compressing is the main reason for these artifacts, which happens because of patch generation and converting to 2D by V-PCC. To compensate for data loss, which is a result of 2D projection, V-PCC has defined more layers of 2D projection across all generated patches to capture more data [1,3]. However, extra layers then increase frame rate, bit rate, and memory buffer, making the method more complex.

To overcome this problem, the proposed method introduces overlapping slicing as an alternative to patch generation to decrease the number of patches generated and the amount of data lost. With the proposed method, bitrate saving is achieved by carefully allocating more layers only where the complexity is greater within the point cloud and where there are increased points of detail to capture. Therefore, the proposed method is much more flexible than the standard V-PCC method.

**Contribution:**

This paper focuses on increasing the compression ratio and decreasing data loss to attain better reconstructed dynamic point clouds quality. To achieve these goals, there are the following contributions:

- Introducing overlapping slice-wise segmentation as an alternative way for patch generation to be able to capture more data and reduce data loss.

- Using slicing to decrease the number of bits required to identify each point's geometric position in the slice results in a greater compression rate while preserving the quality of the reconstructed point cloud.

- Separating slices from where the point cloud's complexity is higher and consequently needs extra layers of 2D projections to capture the points accurately. This helps avoid self-occlusion during 2D projection and keeps more of the original data.

## II. LITERATURE REVIEW

Although MPEG finalized the latest standardization in 2020 and early 2021which are known as G-PCC and V-PCC, there are some limitations for both, which show there is room for further improvement of existing technologies. Geometry-based compression encodes the content directly in the 3D space, while the V-PCC coding is based on converting 3D to 2D data for compression and then back again. G-PCC utilizes data structures and their proximity, such as an Octree that describes the location of the points in the 3D space, whereas the other main compression technology, V-PCC takes advantage of currently available 2D video compression and converts the 3D point data to the collection of multiple 2D images [16-18].

Extending the performance of G-PCC to the temporal axis and using G-PCC for a dynamic point cloud is problematic. First, in the 3D space, each point has 26 neighbours instead of the 8 in the 2D space. There are many possibilities for moving neighbouring points. Still, many of them are transparent because the point cloud has very sparse occupancy of the 3D space, and usually, an object is represented by its surface, not by its volume. Therefore, the prediction of movement using G-PCC is difficult. The second problem with using G-PCC in dynamic point cloud compression is that there is often a significant change in the Octree structure even after a slight movement of the object [1, 3]. The reason is that leaves can jump from one branch of the Octree to another, leading to a significant change in the Octree structure. Therefore Video-based Point Cloud Compression (V-PCC) is more suitable for dynamic point cloud sequences due to its better performance than G-PCC [1,2].

V-PCC decomposes an input point into a set of patches which can then be independently mapped and packed to be encoded by existing video compression such as High-Efficiency Video Coding (HEVC) [19,20]. However, applying 2D video coding on frames that include many converted 3D to 2D patches with unused space is problematic, resulting in the inefficiency of video compression. Therefore, researchers are trying to improve video coding efficiency by using different methods such as cuboid partitioning [21,22] and rate control [23,24]. Researchers have also been attempting to improve the patch generation process and make it more suitable to compress converted 3D data. These efforts include working on unoccupied pixels between patches [25-26] and applying 3D motion estimation [27,28], e.g., using edge detection for orienting motion [29].

(A. Costa et al., 2019) [30] proposed a new patch packing method to improve the V-PCC standard's rate-distortion (RD) performance. Several novel patch packing algorithms were explored along with associated sorting and positioning metrics, both absolute and relative. The RD performance gained using A. Costa et al.'s method, in particular, showed notable improvement for both color and geometry but less success for other areas. They changed the order of the different patches in the 2D atlas, which affected the quality of the reconstructed 3D object. (L. Li et al., 2021) [31] have proposed a method for decreasing unoccupied pixels among different patches due to the inefficiency of coding unused space during video compression. L. Li et al. proposed an occupancy map-based RD optimization that improved the compression efficiency but needed more RD performance. (S. Rhyl et al., 2020) [32] proposed their CHD method for obtaining a contextual homogeneity-based patch decomposition because the TMC2 has limitations of having a patch with multiple different contextual regions, affecting compression efficiency. Their method prevents a single patch from having more than one contextual region in terms of colour and geometry. However, it does not work on additional attributes such as reflection and material ID.

(N. Sheikhipour et al., 2019) [33] proposed a method for improving the coding efficiency and reconstruction quality using a technique in the patch generation process and a single layer. This technique found the most critical patches of the far layer and included them in the patch generation of the near layer. Therefore, on the one hand, this method needs less frame rate and memory buffers, leading to being considered less complex compared to the dual-layer process. On the other hand, the included extra patches of data in the first layer help to improve the coding performance compared with considering only a single layer. However, the geometry quality of the reconstructed point cloud is far below the V-PCC standard. (W. Zhu et al., 2021) [34] proposed optimizing the visual experience for the main view of the user by reserving more points for the patches related to a pre-defined main view. (D. Wang et al., 2021) [35] also aimed to optimize the main view for the user by using the points often discarded during patch generation but only for the main view. They kept points only visible from the main view by reserving more points for the main view patches, which improved only the main view but increased the bitrate of compression.

As a result of those issues outlined above, a reconstructed point cloud using V-PCC suffers from diverse artifacts, especially when high quantization parameters (QP) are applied. K. Cao and P. Cosman 2021 [36] categorized various geometric compression artifacts and then introduced an algorithm for detecting and removing those artifacts. Another artifact removal has been proposed by J. Wei et al. using deep learning [37]. In this paper, the reasons for these artifacts appearing are investigated. For those reasons, a pre-processing method is proposed to minimise the possibility of the appearance of these artifacts.

## III. PROPOSED METHOD

As the proposed method prepared a novel pre-processing tool to improve the efficiency of V-PCC, the V-PCC method will be explained first to find where V-PCC loses data and how the proposed method is able to overcome the limitations of V-PCC.

### A. V-PCC Method and its limitations Explained

To be able to take advantage of existing available 2D video compression, the V-PCC method converts 3D point cloud data to 2D, then 2D data is coded by 2D video encoders. Projecting a 3D image onto the faces of a cube can cause a loss of data and convert back from 2D to 3D may generate significant distortion. Therefore, a clustering process called patch generation is needed before projection. While projecting 3D patches onto 2D images, V-PCC introduces three projection maps to cover all the dimensions of the 3D point cloud. The first map is the geometry map which can be created by embedding depth values. The second map is the texture map which can be created by inserting the related attribute information, e.g., color, light. The third map is the occupancy map which shows which area of the 2D maps is occupied by the points of the point cloud. The three different maps are compressed via 2D video compression to create a bitstream.

### B. Reasons for losing data in V-PCC

With V-PCC, some points are missed, which can cause a hole or crack in the reconstructed point cloud resulting in a loss of quality and detail that are consequences of two issues: patch generating and projecting to 2D.

*1) Losing data through the patch generation process*

The number of patches depends on how complex the point cloud is; usually, this is around a hundred patches for about 95% of the points of a point cloud [1]. The remaining points not covered in this patching process are only isolated points in sparse areas of the point cloud, which would need many patches to capture them. Thus, the encoder may disregard these points as they do not have enough room for many more patches. For this reason, they may introduce some groups of isolated points which are treated differently, either in special patches or as row data [1]. V-PCC may ignore isolated points because of the need for many more patches for isolated areas, and therefore there may be missing disregarded points. In addition, data loss always appears around the edges of patches.

*2) Losing data through converting to 2D data*

The other reason for losing data is self-occlusion while projecting to 2D. Self-occlusion occurs when there is more than one point with the exact absent coordinates of dimension (3rd dimension), which means the points are obscured when projecting a 3D object onto 2D. Consequently, self-occluded points cannot be appropriately captured, as shown in Fig. 1 in purple points, while the green dots are accurately preserved points. Fig.1 shows that some areas around longdress's dress folds and body contours are more at risk of data loss after converting 3D → 2D → 3D. Although V-PCC has defined more than one layer, even up to 16 layers for each geometry and texture map, the problem of lost points has not been solved because of self-occlusion. This occurs because V-PCC applies its fixed layers of the maps right across the point cloud and uses patches projected on different planes. However, as shown from the images in Fig.1, some areas need more layers of projecting onto 2D maps than others because of the higher detailed complexity of those areas.

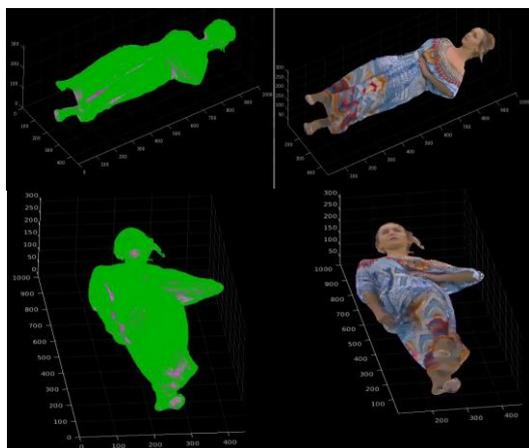

Fig.1. The pink points on the left images are more at risk of losing during 2D projection.

### C. Data loss consequences in reconstruction

V-PCC loses some points due to ignoring isolated points and projecting patches on the 2D plane, which results in about a 10% loss of points. Two types of geometric artifacts are outliers and cracks [36], which occurred because of missing some points. Points were missed because of the many generated patches by V-PCC and 2D projection.

Most artifacts occurred at the edge of patches. Segmenting a point cloud into patches and then separating them for 2D projection results in separating those points that were neighbors in the original point cloud but ended up at the edges of different patches. These neighboring points may be projected onto different planes, and then lossy compression will be applied to these projected patches. Therefore, the edges of patches might not align exactly in the reconstructed point cloud resulting in some cracks. Fig. 2 shows some artifacts appearing after V-PPC compression on the reconstructed 3D point cloud. The artifacts, including cracks (Fig. 2a and 2b) and outliers (Fig. 2c), are mainly raised due to data loss. Fig 2a illustrates some edge cracks that occur at the edges of patches because of separating adjacent points. The other kind of lost points is shown in Fig 2. b, which are the consequence of self-occlusion in 2D projection. Fig 2.c displays outliers resulting from lossy compression around the edge of patches.

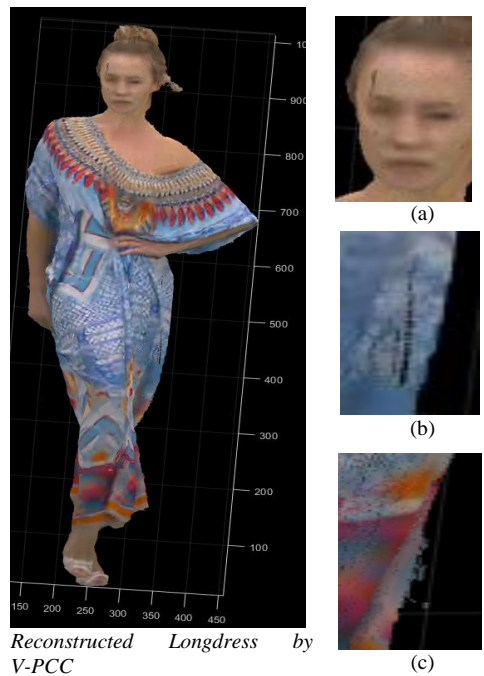

Reconstructed Longdress by V-PCC

Fig.2. Different artifacts caused by V-PCC. (a) edge cracks, (b) projection cracks, and (c) outliers.

### D. The proposed solution to minimise data lost with V-PCC

As the search for self-occluded points continued in assessing the efficiency of the current process of projecting dynamic point clouds onto 2D planes, it was revealed that there was a severe lack of points available in 2D projection, especially in the areas of, for example, folds in clothes/material and where there are contours of a human or animal body or statue, which are demonstrated in Fig. 3. In Fig. 3, the green points are those points that can be successfully captured by the first layer of 2D projection. However, the rest are the points that cannot be captured using the first layer since they have different angles from the projection plane and, therefore, will self-occlude after projection.

As mentioned above, to overcome the problem of self-occlusion in 2D projection, V-PCC introduced a multi-layer approach in which the point cloud information is projected onto multiple layers of 2D maps. Extra layers have been added to contain those points that cannot be mapped into the first layer. However, it is sometimes a waste of valuable bitrate to have multiple layers across the whole point cloud since some areas of any point cloud have very sparse points and a lower possibility of self-occlusion. In addition, in each of these layers of maps, there are several patches of the different 2D projections leading to losing data around the edges of the patches. However, increasing the maps across the point cloud evenly (in the V-PCC method) is unnecessary because the number of occluded points will not be the same over the whole point cloud. To minimise data loss, a novel pre-processing method is introduced to cross-section overlapped slices of the point cloud more intensely, but only in complex areas where more self-occluded points are occurring.

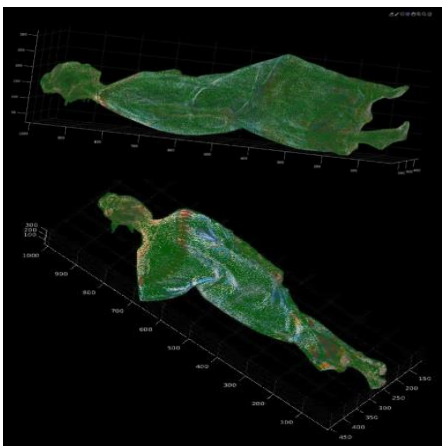

Fig.3. Green points are captured successfully, but the rest are missed using the first layer of 2D projection.

*E. The proposed method overview*

To address the issues outlined above and in the literature review, a new approach to pre-processing the point cloud is proposed, utilizing the best parts of the existing standard method and introducing targeted slicing. This can reduce data loss due to self-occlusion and generated patches. Therefore, more available points can be captured to increase the quality and, at the same time, reduce the amount of data needing compression. The proposed targeted slicing of the point cloud in pre-processing adds layers only where required to minimise self-occlusion and decrease bitrate. Therefore, instead of having more fixed 2D maps for the whole point cloud and sacrificing bitrate (as in V-PCC), the proposed method adds extra layers only in the complex areas. The number of layers of the map can be changed, giving added flexibility in capturing all points.

*F. The proposed method in more detail*

Firstly, the areas of the point cloud which are more complex and have more layers or contours of data to capture should be found. These areas would need increased maps for converting to 2D because they have more risk of self-occlusion. Here, targeted slicing of the original point cloud will be introduced to preserve the detail so that after 2D projection, more detailed data can be restored. The point cloud is sliced around the whole point cloud initially. The maximum width of each slice is defined so that the slice can be identified using reduced bits. Knowing the position of the starting point for each slice with the assistance of geometry differences can lead to a decrease in the number of bits required to determine the position of each point in the slice. The proposed method will use slicing to reduce the 10 bits of geometry (depth position) to 6 bits because the "θ" value is defined as 64. Each slice is treated as a new point cloud, with a reduced number of bits, to identify the position of all points of the point cloud slice.

*1) How to reduce the number of bits, using differences*

The position differences can be applied to locate the points according to the value of differences from the beginning point of the slice. In other words, a position of a point concerning a base of a slice can be calculated using geometry differences; consequently, the position of the point in the point cloud can be achievable. For each slice, some bits will be allocated according to the width of the slice to find the geometry value. For example, if the slice starts from 220 – 250 along the z-axis, the difference would be 30 units. Using the difference and knowing the base for the whole slice is 220 can help reduce the number of bits of data information needing to be sent. E.g., if the point is supposed to be noted as 228 on the z-axis, instead of sending 228, it may be possible to send just 8 (using fewer bits) since it should be added to 220 as the base. For example, a few consecutive slices of Loot along the direction of the z-axis are shown in Fig. 4.

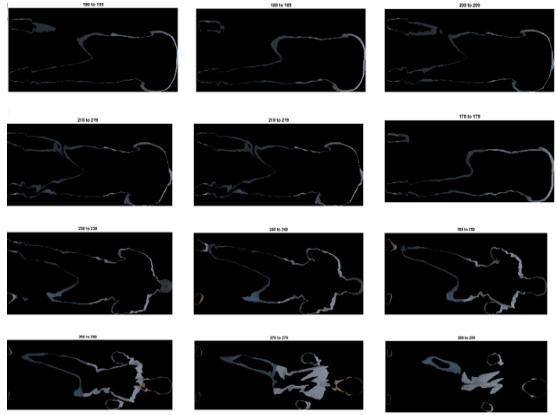

Fig.4. Slicing Loot point cloud sequence only in one direction

A maximum width size " θ " for slices should be pre-defined first. Then according to this maximum, the number of bits required for each slice is calculated. E.g., if the maximum width is 64, then only 6 bits are needed to identify the slice instead of 10 bits. This saves the required bitrate. A threshold value will also be defined to distinguish whether a slice is going to be a new point cloud or whether it should be sliced from a different direction to include more points of the total point cloud.

*2) How to slice the point cloud*

Since the aim is to capture more data in each slice, firstly, the exact size of each slice and secondly, the best direction of slicing should be defined according to the slightest possibility of missed data. To this end, the following steps are introduced:

a. If one slice has less than a threshold number of points, slicing is stopped from that direction because there are not enough points to require further slicing. This threshold number of points depends on whether there is a need to slice from the other perspectives to include more points in that one slice. The proposed method defines the threshold value as 5% of the total points in the point cloud.

b. The width of slices can vary up to the defined maximum, depending on the number of points at risk of being lost. To determine the size, the ends of each side of the point cloud are sliced individually, from up to the maximum width of "θ," so that the collection of 2D projections of this slice can cover maximum points; therefore, less occlusion occurs after 2D projection. This can be calculated using the following formula:

$$\psi_P = (\phi_P - \sum_{k=1}^{n} \alpha_k )/\phi_P \qquad (1)$$

where "$\psi_P$" is the number of lost points after 2D projection and "$\phi_P$" is the points in the new sliced point cloud, and "$\alpha_k$" is the area of the 2D projection of the $k_{th}$ connected component of the slice on the planes. According to the priority of minimising occluded points and with the assistance of this formula, the exact width of each slice can be found in any sizes less than "θ" which prproduceshe least amount of "$\psi_P$" and is selected as the size of the slice in that direction.

c. Slicing is done from around the point cloud. As seen in Fig. 5, slicing only from one direction creates a hole in the 2D projection.

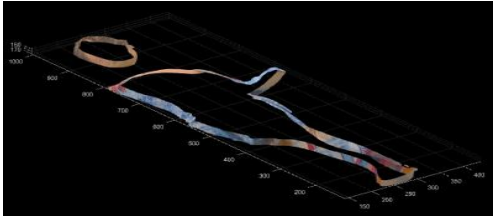

Fig.5. One slice from the middle of Longdress.

d. After slicing from each of the six sides (±x, ±y, ±z), all those selected slices will be compared with each other using the above formula (1). Any slice that includes less data loss is selected as a newly selected slice, and the rest of the point cloud will be considered a new point cloud to be able to slice it again and capture as many points as possible. This equation reveals the number of points that are lost as a result of self-occlusion. Therefore, it helps to find which slice width in which direction (±x, ±y, ±z) has minor self-occluded points, and then this slice will be individually selected and treated as a new point cloud.

e. The above steps will be repeated for the rest of the point cloud until all points in the point cloud are considered, and there is no need to slice any further to capture the points. From this step, then slices are compressed using V-PCC.

f. The above steps will be repeated for the rest of the point cloud until all points in the point cloud are considered, and there is no need to slice any further to capture the points. From this step, then slices are compressed using V-PCC.

Because of not slicing just from one direction, the possibility of creating a hole in the selected slice is lessened, increasing the efficiency of video compressing (see Fig. 4 and Fig. 5),. This is also shown in Fig. 6, which includes six slices from 6 sides of Longdress, Redandblack, and Loot slicing.

g. Some distortions around each slice are visible after compression, which may result in losing points. It is proposed to overlap slices for a couple of lines around each slice to maintain those points.

Note: Mostly, after one repetition of slicing around the whole point cloud, the rest of the slicing will not be along the longest axis because the slices do not include as many points as needed to be less than the threshold. Since covering more available points is the target of this slicing method, more points are captured in each slice.

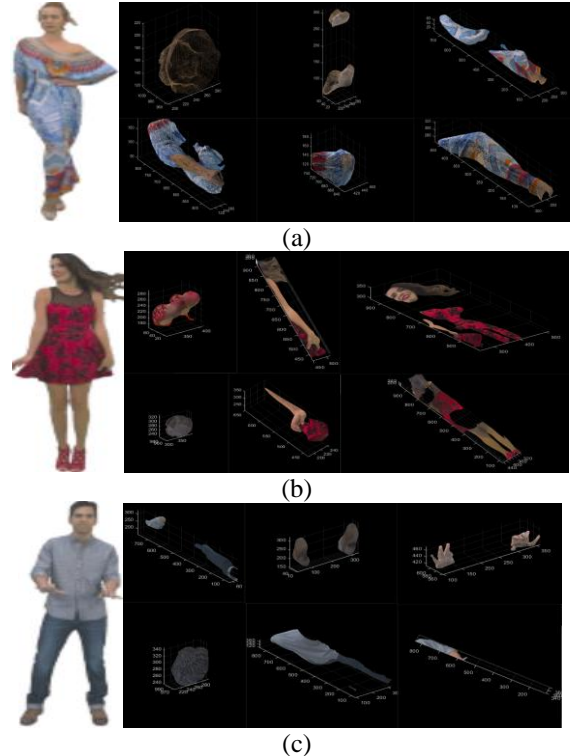

Fig.6. Point cloud slicing around 6(±x, ±y, ±z) directions shows the connected components need less space for 2D projection than Fig.5. (a)Longdress, (b)Redandblack, and (c)Loot

## IV. EXPERIMENTAL RESULTS

The performance of the proposed method in this paper was determined using V-PCC reference software TMC2 to compare their performance. Longdress, Loot, and Redandblack have been chosen as standard test dynamic point clouds. There is a visual comparison to observe in addition to objective quality evaluations. Fig. 7 shows the original point cloud, and its reconstruction via V-PCC, and the proposed method. Fig. 7 is enlarged in three steps to show the differences more clearly. As can be observed, there is a noticeable crack on the face of Longdress using V-PCC, as well as a disfigurement appearing at her neck where her hair had a curl in the original, and these are both improved using the proposed method.

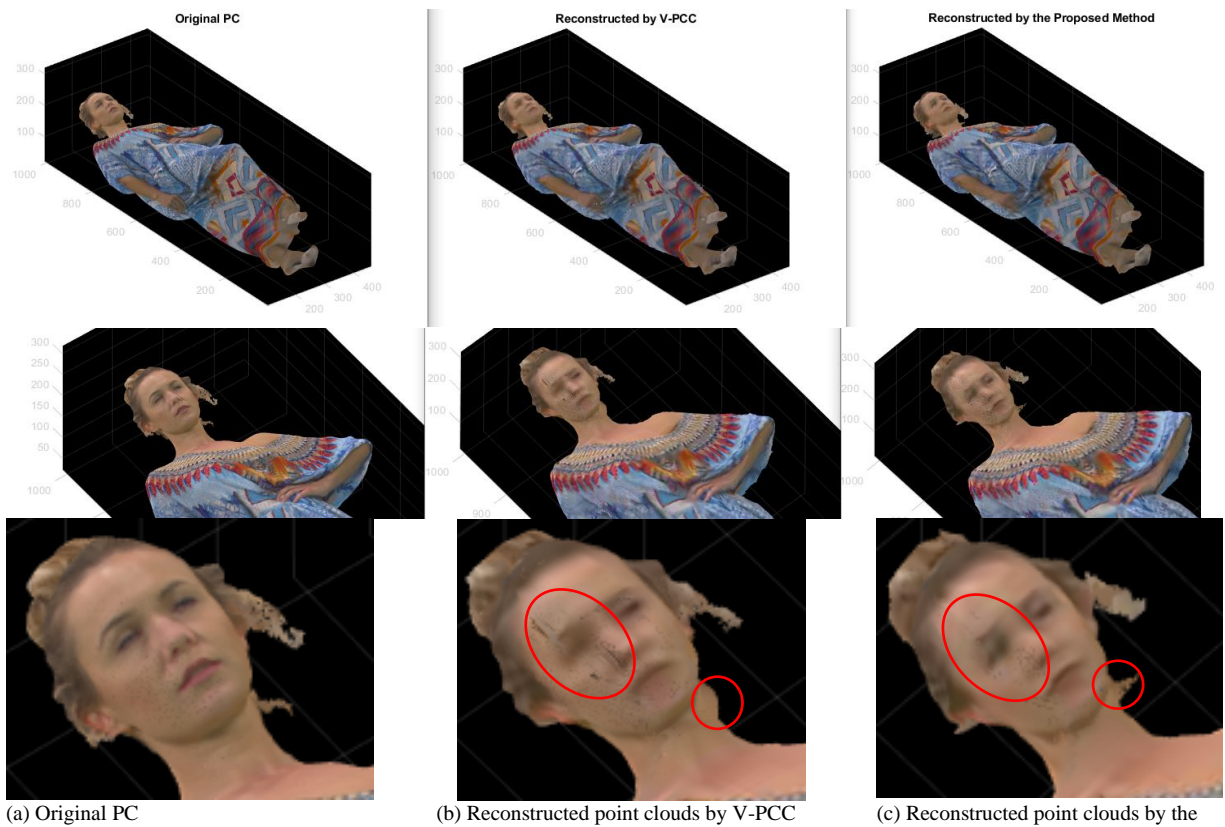

(a) Original PC  (b) Reconstructed point clouds by V-PCC  (c) Reconstructed point clouds by the proposed method

Fig.7. Comparing reconstruction of the compressed point cloud using V-PCC and the proposed method (a) the original point cloud, (b)3D reconstructed point cloud using V-PCC, and (c) 3D reconstructed point cloud using the proposed method

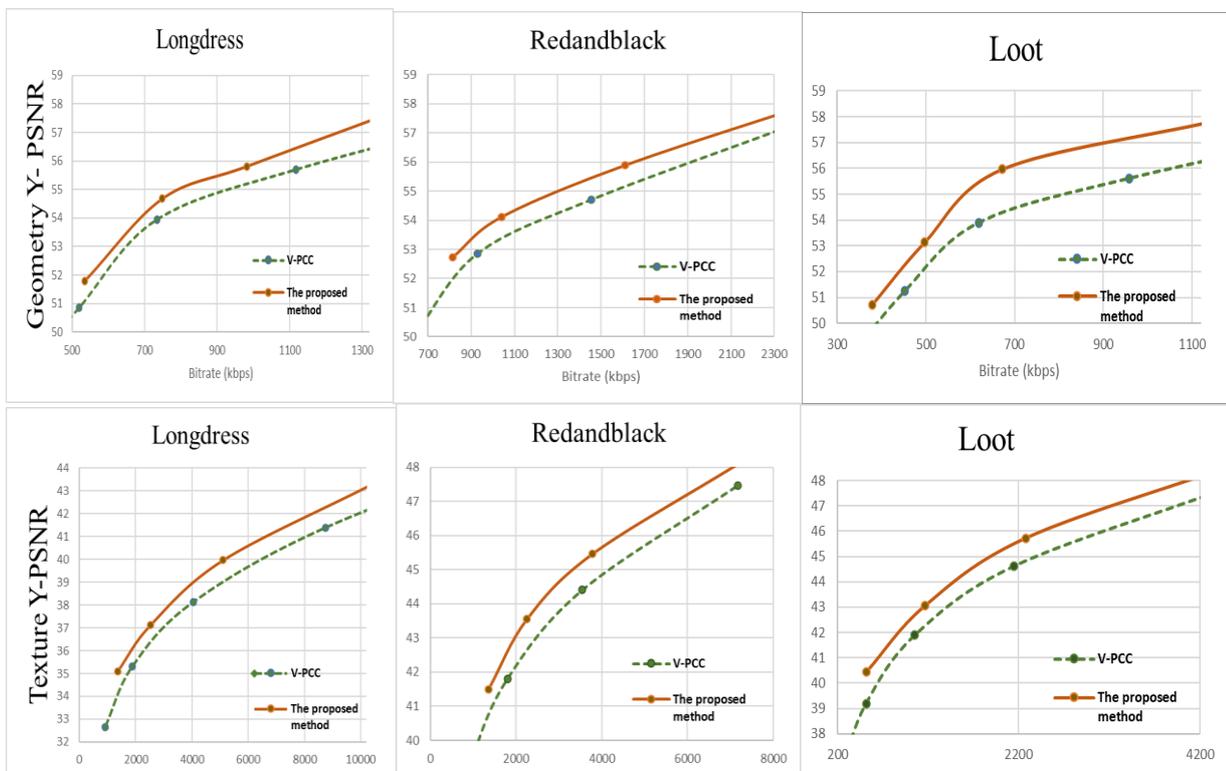

Fig.8. The rate-distortion curves by the proposed method (orange-solid) and the V-PCC standard method (green-dotted) using different standard video sequences.

The crack appearing is due to the patch method causing data loss when two patches for the face are used in V-PCC and projected onto different planes, but both patches have lost their connection and proximity. The other reason for cracks appearing is data is lost at the edges of patches. There are three reasons for improving cracks using the proposed method. 1. Decreasing the number of self-occluded points, 2. Projecting adjacent points onto the same plane, and 3. Overlapping slices to retain more original points, overcoming the loss of original data and its connection to one another, and minimising cracks. Ignoring isolated points by V-PCC causes the disfigurement to appear at the neck, which is also improved using the proposed method because the proposed method can include more isolated points. In other words, using Slice-Wise Overlapping Segmentation as a pre-processing by the proposed method helps to maintain more original data resulting in fewer artifacts appearing in the reconstructed point cloud.

TABLE 1: Comparison of data loss between the proposed and V-PCC methods for different types of datasets.

| Sequence | Data loss by V-PCC | Data loss by the proposed method |
|---|---|---|
| Redandblack | 9.4% | 8% |
| Loot | 12% | 10.4% |
| Longdress | 9.8% | 8.3% |
| Average | 10.4% | 8.9% |

TABLE 2: BD-Bit rate and BD-PSNR of different types of videosequences of Geometry performance.

| Sequence | BD-Bit rate | BD-PSNR |
|---|---|---|
| Redandblack | -14.4% | 0.85 |
| Loot | -15.3% | 1.25 |
| Longdress | -8.4% | 0.9 |
| Average | -12.7% | 1 |

TABLE 3: BD-Bit rate and BD-PSNR of different types of video sequences of Texture performance.

| Sequence | BD-Bit rate | BD-PSNR |
|---|---|---|
| Redandblack | -20.3% | 0.9 |
| Loot | -18.5% | 0.9 |
| Longdress | -19% | 0.75 |
| Average | -19.3% | 0.85 |

The RD curves using the proposed method when θ = 64 (orange-solid) and the V-PCC standard method (green-dotted) of different data sets are displayed in Fig. 8., including Redandblack, Loot, and Longdress video sequences. These experimental results prove that the proposed method improves the RD performance and increases the quality of reconstructed 2D maps of the point cloud compared to the standard V-PCC method. Fig.9. shows a comparison of three different areas of the reconstructed point cloud of Longdress using V-PCC on the left and the proposed method on the right. In Fig.9, on the top is a portion of her dress fold, including the green points that are appropriately reconstructed. As it can be seen, the green points are more using the proposed method shown on the right hand because the points were captured more accurately. The middle image shows the fold of her dress from another viewpoint, the pink areas show those points that are different from the original ones, and the green areas are correctly reconstructed. The image on the bottom of Fig. 9 includes another fold of Longdress' dress in a different slice, and once again, the number of green points is demonstrably higher using the proposed method. Overall, the total number of correctly captured and reconstructed points are shown to be at least 1.5 % more using the proposed method. The amount of data loss is listed in Table. 1, shows that the proposed method is able to capture more data more accurately, and therefore, the data lost with the proposed method is less. The BD-Bit rate and BD-PSNR for these four different standard video sequences are shown in Table 2 and Table 3.

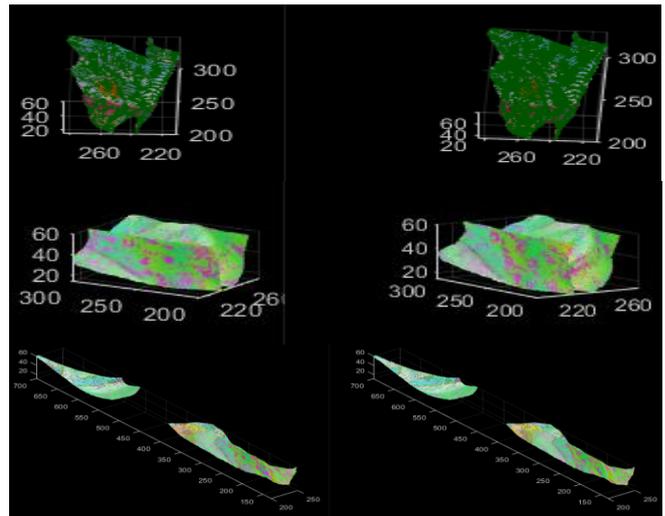

Fig.9. Comparing data lost in some critical areas of the reconstructed point cloud Longdress using V-PCC (left) and the proposed method (right). The original data are in green, and the rest are artificial because of losing data.

V. CONCLUSION

In this paper, the limitations of V-PCC have been analysed to find where V-PCC is efficient and where it can be improved. The reconstructed point cloud using V-PCC suffers from artifacts and limitations, mainly due to losing original points when captured data is being processed. Patch generation and converting 3D/2D were found to be the most concerning areas to address. To minimise the artifacts that occur using V-PCC, this paper proposed a new method of pre-processing a dynamic point cloud using overlapping slice-wise segmentation to capture and preserve more original data, improving the quality of the reconstructed point cloud. The size of each slice is defined according to the areas of complexity in each specific point cloud and therefore is more flexible than V-PCC layers.

The proposed method using differences to address the geometric position of each point has shown a decrease in the required bitrate in higher quality. The experimental results show improved quality of the reconstructed point cloud and less occurrence of artifacts and cracks using the slice-wise overlapping segmentation pre-processing method. This will improve the efficiency of V-PCC during pre-processing because it enables the capture of more original data and, therefore, reconstruction of a more realistic view of the original point clouds, which has been proved in the experimental results.